\documentclass[11pt]{article}
\usepackage[a4paper,margin=1in]{geometry}
\usepackage[T1]{fontenc}
\usepackage{newtxtext,newtxmath}
\usepackage{microtype}
\usepackage{amsmath}
\usepackage{graphicx}
\usepackage{booktabs}
\usepackage{multirow}
\usepackage{array}
\usepackage{tabularx}
\usepackage{makecell}
\usepackage{float}
\usepackage{caption}
\usepackage{pdflscape}
\usepackage[colorlinks=true,linkcolor=blue,citecolor=blue,urlcolor=blue]{hyperref}

\begin{document}

\thispagestyle{empty}
\vspace*{-0.5in}

\noindent\rule{\textwidth}{1.2pt}
\vspace{0.4in}

\begin{center}
{\Large\bfseries
Deep Convolutional Architectures for EEG Classification:\\
A Comparative Study with Temporal Augmentation and\\
Confidence-Based Voting
}
\end{center}

\vspace{0.4in}
\noindent\rule{\textwidth}{1.2pt}

\vspace{0.6in}

\noindent
\begin{minipage}[t]{0.48\textwidth}
\begin{center}
{\bfseries Aryan Patodiya}\\[4pt]
Department of Computer Science\\
California State University, Fresno\\
Fresno, CA, USA\\
\texttt{apatodiya@mail.fresnostate.edu}
\end{center}
\end{minipage}
\hfill
\begin{minipage}[t]{0.48\textwidth}
\begin{center}
{\bfseries Hubert Cecotti}\\[4pt]
Department of Computer Science\\
California State University, Fresno\\
Fresno, CA, USA\\
\texttt{hcecotti@mail.fresnostate.edu}
\end{center}
\end{minipage}

\vspace{0.7in}

\begin{center}
{\Large\bfseries ABSTRACT}
\end{center}

\vspace{0.25in}

\noindent
Electroencephalography (EEG) classification plays a key role in brain-computer interface (BCI) systems, yet it remains challenging due to the low signal-to-noise ratio, temporal variability of neural responses, and limited data availability. In this paper, we present a comparative study of deep learning architectures for classifying event-related potentials (ERPs) in EEG signals. The preprocessing pipeline includes bandpass filtering, spatial filtering, and normalization. We design and compare three main pipelines: a 2D convolutional neural network (CNN) using Common Spatial Pattern (CSP), a second 2D CNN trained directly on raw data for a fair comparison, and a 3D CNN that jointly models spatiotemporal representations. To address ERP latency variations, we introduce a temporal shift augmentation strategy during training. At inference time, we employ a confidence-based test-time voting mechanism to improve prediction stability across shifted trials. An experimental evaluation on a stratified five-fold cross-validation protocol demonstrates that while CSP provides a benefit to the 2D architecture, the proposed 3D CNN significantly outperforms both 2D variants in terms of AUC and balanced accuracy. These findings highlight the effectiveness of temporal-aware architectures and augmentation strategies for robust EEG signal classification.

\vspace{0.4in}


\vspace{0.4in}

\section{Introduction}
\label{sec:introduction}

Electroencephalography (EEG) is a widely used non-invasive neuroimaging modality that records electrical activity from the scalp with millisecond-level temporal resolution ~\cite{niedermeyer2005electroencephalography}. EEG-based classification plays a fundamental role in non-invasive brain-computer interface (BCI) systems, particularly in detecting event-related potentials (ERPs), which are time-locked neural responses to external stimuli~\cite{luck2014introduction}. Accurate and robust classification of ERPs can facilitate a range of applications, including attention monitoring, communication interfaces, and neurorehabilitation~\cite{wolpaw2002brain}.

Despite its potential, single-trial EEG classification remains challenging due to the inherently low signal-to-noise ratio, the presence of artifacts, and inter-session and inter-subject variability~\cite{lotte2018review}. The temporal latency of ERP responses can fluctuate across trials, introducing additional challenges in fixed-window analysis ~\cite{crone2006neurophysiological}. These factors necessitate the development of models that are both spatially and temporally adaptive.

In recent years, deep learning methods have shown strong performance in EEG signal classification, often outperforming traditional machine learning pipelines~\cite{bashivan2015learning} ~\cite{schirrmeister2017deep}. To enhance spatial discriminability, methods, e.g., Common Spatial Pattern (CSP), have been employed as a preprocessing step ~\cite{blankertz2008optimizing}. Combined with convolutional neural networks (CNNs), this enables the network to learn high-level temporal and spatial features from the projected EEG signals~\cite{lawhern2018eegnet}.

In this work, we present a comprehensive evaluation of two deep architectures for EEG classification: a 2D CNN and a 3D CNN. To ensure a fair comparison and evaluate the impact of preprocessing, we test the 2D CNN in two configurations: one using CSP for feature extraction, and another trained directly on raw sensor data. The 2D CNN is structured to learn temporal and spatial representations through sequential convolutional blocks, while the 3D CNN jointly captures spatiotemporal patterns using volumetric EEG input. Inspired by prior work on 3D CNNs for ERP detection~\cite{Cecotti2019_3DCNN} and the use of artificial trials for handling temporal variability~\cite{Cecotti2015_ArtificialTrials}, we apply temporal shift augmentation during training and introduce a confidence-based voting mechanism at test time across multiple shifted versions of each trial.

Our contributions can be summarized as follows: 1) We design and compare three distinct pipelines: a 2D CNN with CSP, a 2D CNN without CSP, and a 3D CNN. 2) We analyze the impact of CSP as a preprocessing step for the 2D architecture. 3) We introduce temporal shift augmentation and a confidence-based test-time voting strategy to address ERP latency variability. 4) We conduct a thorough empirical evaluation demonstrating the superior robustness and generalization of the 3D CNN architecture in a fair comparison.

The rest of the paper is organized as follows. The experimental protocol for acquiring data and the different neural network architectures are presented in Section~\ref{sec:methods}. The performance of the various classifiers is given in Section~\ref{sec:results}. Finally, the results are discussed in Section~\ref{sec:discussion}, with the main findings summarized in Section~\ref{sec:conclusion}.

\section{Methods}
\label{sec:methods}

\subsection{Experimental protocol}

Healthy adults participated in a rapid serial visual presentation task where color images of faces (front-facing) were presented to the subjects. Images contained targets (male face) and non-targets (female face). The goal was to count the number of target images presented mentally. The study was approved by the institutional review board for human subject research at California State University, Fresno. The duration between two stimuli was about 500~ms. There was no inter-stimulus interval. The target probability was set to 10\% in blocks of 20 images. For each participant, there was a total of 80 targets and 720 non-target images.

\begin{figure}[htbp]
    \centering
    \begin{tabular}{cccccccc}
        \includegraphics[height=1.4cm]{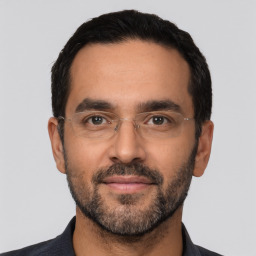} &
        \includegraphics[height=1.4cm]{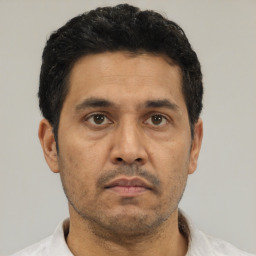} &
        \includegraphics[height=1.4cm]{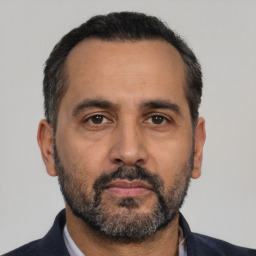} &
        \includegraphics[height=1.4cm]{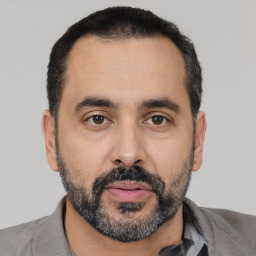} &
        \includegraphics[height=1.4cm]{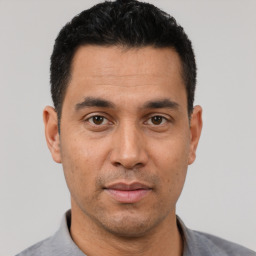} &
        \includegraphics[height=1.4cm]{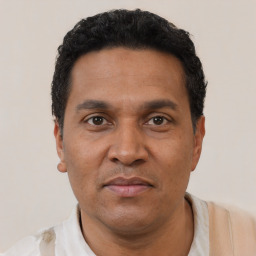} & 
        \includegraphics[height=1.4cm]{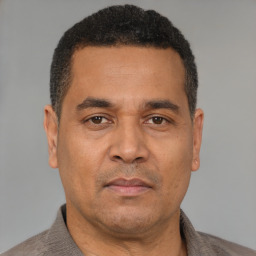} &
        \includegraphics[height=1.4cm]{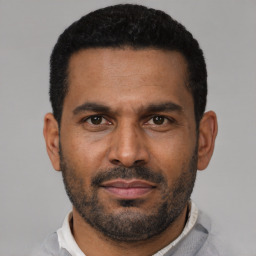}\\
        \includegraphics[height=1.4cm]{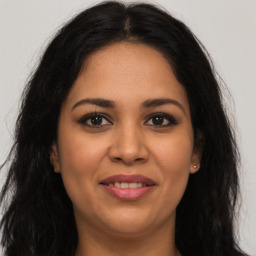} &
        \includegraphics[height=1.4cm]{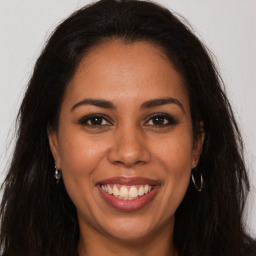} &
        \includegraphics[height=1.4cm]{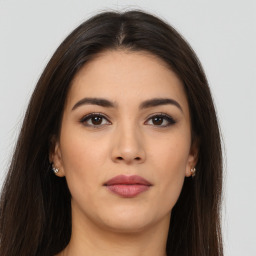} &
        \includegraphics[height=1.4cm]{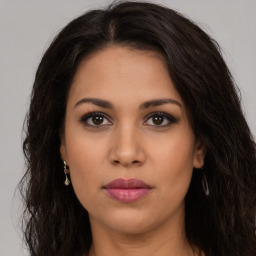} &
        \includegraphics[height=1.4cm]{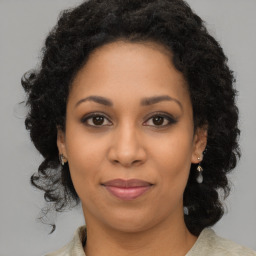} &
        \includegraphics[height=1.4cm]{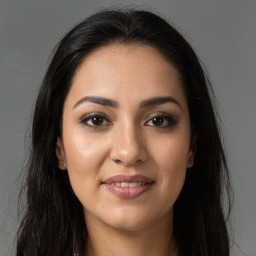} & 
        \includegraphics[height=1.4cm]{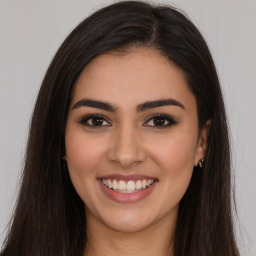} &
        \includegraphics[height=1.4cm]{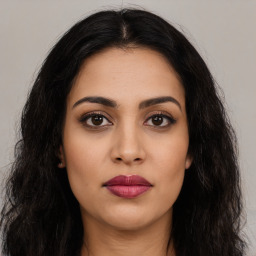}
    \end{tabular}
    \caption{Images of targets (top) and non-target (bottom).}
    \label{fig:stimuli}
\end{figure}

EEG data were recorded using a Biosemi ActiveTwo system with a 32-channel electrode cap following the international 10–20 system. For both the 2D and 3D CNN pipelines, we used all 32 scalp electrodes without any channel reduction. This full-channel configuration preserves spatial richness and enables the models to learn from complete topographical brain activity.

Each epoch was assigned a binary label based on its associated trigger code. Target events were labeled by trigger code 115 (label 1) and non-target events by code 234 (label 0). Trials with unknown or missing triggers were discarded.

\subsection{Signal processing}

The first step involved reading the raw BDF files using a custom MATLAB parser. All 32 EEG channels were extracted, and trials were segmented using the annotated trigger values. Each trial was bandpass filtered between 0.1~Hz and 30~Hz using a 4th-order Butterworth filter to remove low-frequency drift and high-frequency noise.

Trials were segmented into fixed-length epochs starting from stimulus onset. Each epoch spanned a 1000~ms window after the stimulus. These were then downsampled to 64 time points by averaging over non-overlapping windows, preserving coarse temporal information while reducing computation.

Following these initial steps, the preprocessing diverged based on the specific model. For the \textbf{2D CNN pipeline with CSP}, two spatial filtering steps were applied. First, a surface Laplacian filter enhanced local activity. Following this, Common Spatial Pattern (CSP) transformation was used to extract six discriminative components, resulting in an input shape of $6 \times 64 \times 1$ for each trial.

To provide a fair comparison against the 3D architecture, a \textbf{second 2D CNN pipeline was evaluated without CSP}. For this approach, the epoched and downsampled data from all 32 sensor channels were used directly as input to the network, with no spatial filtering applied during preprocessing. The input shape was $32 \times64 \times 1$ for each trial.

\subsubsection{Mapping Sensor Coordinates from 2D to 3D Grid Layout (3D CNN Only)}

To represent spatial information for the 3D CNN, we approximated the electrode positions using a $7 \times 5$ 2D spatial grid reflecting the standard 10–20 layout~\cite{silverman1963rationale}. 28 out of the 32 scalp electrodes were assigned to known grid positions based on their anatomical locations. For grid cells without a direct electrode assignment, we applied spatial interpolation using the nearest neighboring electrodes to estimate their values. This resulted in a smooth and spatially coherent volume for each time point, minimizing discontinuities and preserving local spatial gradients.

\begin{table}[htbp]
\centering
\caption{2D sensor layout.}
\begin{tabular}{|c|c|c|c|c|}
\hline
F7& F3& Fz& F4& F8\\
FC5& FC1 & - & FC2& FC6\\
T7 & C3  & Cz  & C4 & T8\\
CP5& CP1& - &CP2 &CP6\\
P7 & P3& Pz& P4& P8 \\
- &PO3& - & PO4 & - \\
- & O1 & Oz &  O2 & - \\
\hline
\end{tabular}
\end{table}

The temporal shift augmentation was applied after this grid mapping. For compatibility with our 3D CNN architecture, each interpolated $6 \times 6$ grid was flattened back into a pseudo-channel axis, resulting in a tensor of shape $N_s \times N_t \times D \times 1 \times N$, where: 1) $N_s=32$ is the number of EEG channels (flattened from the 2D grid). 2) $N_t=64$ is the number of time points. 3) $D=5$: number of temporal shifts used (-2, -1, 0, 1, 2. 4). 4) 1: singleton channel dimension. 5) $N$ is the number of trials.

\subsubsection{Temporal Shift Augmentation (3D CNN Only)}

To account for ERP latency variability, we used temporal shift augmentation by generating four additional shifted versions of each trial at offsets $\Delta t = \{-2, -1, 0, +1, +2\}$. These temporally jittered copies were stacked along a new depth axis. This augmentation makes the model more robust to variations in the timing of neural responses. The resulting 5D tensor has the shape $N_s \times N_t \times D \times 1 \times N$ for each trial.

\subsubsection{Data Normalization}

We applied per-trial z-score normalization to standardize the signal distributions. For each trial, the mean and standard deviation were computed across all channels and time points, and the data were normalized accordingly. This process avoids data leakage and stabilizes training. We employed stratified 5-fold cross-validation to ensure class balance across folds. Each fold had disjoint train, validation, and test sets. The preprocessed data for each fold were saved independently for reproducibility and consistency during model evaluation.

\subsection{Classifiers}

This section describes the various deep learning architectures developed and evaluated for ERP-based EEG classification. We implemented and tested multiple variants of two primary architectures: a two-dimensional Convolutional Neural Network (2D CNN) and a three-dimensional Convolutional Neural Network (3D CNN). Each model type was designed to operate on a specific representation of EEG trials, to learn robust spatial and temporal patterns for classification.

The 2D CNN models were evaluated in two distinct configurations to assess the impact of preprocessing. The first set of models was applied to EEG signals that had undergone preprocessing and spatial projection using CSP. These spatially filtered trials were reshaped into 2D input matrices for the network. The second set of models, intended for a fair comparison with the 3D architecture, was trained directly on the raw 32-channel sensor data without CSP. For both configurations, we experimented with different 2D CNN variants, varying activation functions (ReLU, Swish, GELU), loss functions (cross-entropy vs. focal loss), and regularization strategies. We also evaluated the impact of architectural modifications such as global average pooling and Squeeze-and-Excitation (SE) blocks. Results from these variants are compared in later sections.

In contrast, the 3D CNN models were trained directly on raw, preprocessed EEG data. These models processed 5D volumetric input tensors that preserved spatial, temporal, and depth information across trials. We introduced temporal shift augmentation~\cite{Cecotti2015_ArtificialTrials} to increase robustness to ERP latency jitter by including multiple shifted versions of each trial (-2 to +2 time steps). In addition to baseline 3D CNNs, we evaluated augmented versions trained on shifted trials, as well as test-time augmentation (TTA) with confidence-based voting across shifts.

While each model variant differs in structure and training setup, all share several common deep learning components. The Focal Loss is used in most experiments to address class imbalance by down-weighting easy examples and emphasizing harder ones. Batch Normalization is applied after each convolutional layer to improve training stability. The non-linear activations being tested are ReLU, Swish, and GELU to identify the best activation for learning ERP component features. Finally, dropout is applied to mitigate overfitting, especially in deeper configurations.

The following subsections describe the details of each model variant, including input shape, architecture, and design rationale.

\subsubsection{2D CNN Architectures}

In this subsection, we present the series of 2D CNN architectures that were evaluated. To assess the impact of preprocessing, these six architectures were tested under two conditions: once on CSP-transformed EEG trials and a second time on raw 32-channel data. For the CSP-based models, the input tensors have a shape of C×T×1, where C is the number of spatial filters. For the models without CSP, the input shape is 32×T×1, where T is the number of time samples. While the core architectural blocks remain similar—comprising temporal and spatial convolutions followed by feature extraction—the activation functions, loss objectives, and regularization strategies varied across experiments, creating the six main variants described below.

\subsubsection*{Baseline 2D CNN (ReLU + Cross-Entropy)}

The baseline model consists of three convolutional layers: a temporal filter, a spatial filter, and a full 2D convolution to extract joint features. ReLU activations are used after each convolution, and the model is trained using the categorical cross-entropy loss. This configuration reflects standard CNN baselines for EEGNet-style models. This baseline served as a starting point but showed signs of overfitting and poor sensitivity to the minority class, leading to suboptimal balanced accuracy across folds.

\begin{table}[H]
\centering
\caption{Baseline 2D CNN (ReLU + Cross-Entropy)}
\begin{tabular}{llll}
\toprule
\textbf{Layer} & \textbf{Filter/Units} & \textbf{Output Shape} & \textbf{Remarks} \\
\midrule
Input & -- & $C \times T \times 1$ & CSP/Non-CSP input \\
Conv2D & $1 \times 10$, $F=8$ & $C \times T' \times 8$ & Temporal \\
ReLU + BN & -- & same & -- \\
Conv2D & $C \times 1$, $F=16$ & $1 \times T' \times 16$ & Spatial \\
ReLU + BN & -- & same & -- \\
Conv2D & $1 \times 10$, $F=32$ & $1 \times T'' \times 32$ & Full 2D \\
ReLU + BN & -- & same & -- \\
Dropout & $p=0.3$ & same & Regularization \\
Flatten + Dense & 64 & 64 & -- \\
Dense + Softmax & 2 & 2 & Cross-entropy output \\
\bottomrule
\end{tabular}
\end{table}

\subsubsection*{Improved 2D CNN (GELU + Focal Loss)}

To improve robustness to class imbalance and introduce smoother nonlinearities, we replaced ReLU with the Gaussian Error Linear Unit (GELU) activation and used the focal loss function. The GELU activation is defined as:

\begin{equation}
\text{GELU}(x) = \frac{1}{2}x \left(1 + \tanh\left[\sqrt{\frac{2}{\pi}} \left(x + 0.044715x^3\right)\right]\right)
\end{equation}

\textit{This configuration consistently outperformed the baseline in both AUC and balanced accuracy and was used as the foundation for later variants.}

\begin{table}[htb] 
\centering
\caption{Improved 2D CNN (GELU + Focal Loss)}
\begin{tabular}{llll}
\toprule
\textbf{Layer} & \textbf{Filter/Units} & \textbf{Output Shape} & \textbf{Remarks} \\
\midrule
Input & -- & $C \times T \times 1$ & CSP/Non-CSP input \\
Conv2D & $1 \times 10$, $F=8$ & $C \times T' \times 8$ & Temporal \\
GELU + BN & -- & same & -- \\
Conv2D & $C \times 1$, $F=16$ & $1 \times T' \times 16$ & Spatial \\
GELU + BN & -- & same & -- \\
Conv2D & $1 \times 10$, $F=32$ & $1 \times T'' \times 32$ & Full 2D \\
GELU + BN & -- & same & -- \\
Dropout & $p=0.3$ & same & -- \\
Flatten + Dense & 64 & 64 & -- \\
Dense + Softmax & 2 & 2 & Focal loss output \\
\bottomrule
\end{tabular}
\end{table}

\subsubsection*{2D CNN with Global Average Pooling}

\textit{The GAP-based variant slightly improved generalization, particularly in validation performance, and reduced overfitting on smaller datasets.}

\begin{table}[H]
\centering
\caption{2D CNN with Global Average Pooling}
\begin{tabular}{llll}
\toprule
\textbf{Layer} & \textbf{Filter/Units} & \textbf{Output Shape} & \textbf{Remarks} \\
\midrule
Input & -- & $C \times T \times 1$ & CSP/Non-CSP input \\
Conv2D & $1 \times 10$, $F=8$ & $C \times T' \times 8$ & Temporal \\
GELU + BN & -- & same & -- \\
Conv2D & $C \times 1$, $F=16$ & $1 \times T' \times 16$ & Spatial \\
GELU + BN & -- & same & -- \\
Conv2D & $1 \times 10$, $F=32$ & $1 \times T'' \times 32$ & Full 2D \\
GELU + BN & -- & same & -- \\
Dropout & $p=0.3$ & same & -- \\
GlobalAvgPool2D & -- & $1 \times 1 \times 32$ & Replaces FC \\
Dense + Softmax & 2 & 2 & Focal loss output \\
\bottomrule
\end{tabular}
\end{table}

\subsubsection*{2D CNN with Squeeze-and-Excitation (SE) Block}

\textit{The SE-enhanced model achieved marginal gains in AUC and showed better stability in fold-wise performance, indicating the value of attention mechanisms in EEG classification.}

\begin{table}[H]
\centering
\caption{2D CNN with Squeeze-and-Excitation (SE) Block}
\begin{tabular}{|l|p{3.6cm}|l|p{4.8cm}|}
\hline
\textbf{Layer} & \textbf{Details} & \textbf{Output Shape} & \textbf{Remarks} \\
\hline
Input & -- & $C \times T \times 1$ & CSP/Non-CSP input matrix \\
Conv2D & $1 \times 10$, $F=8$ & $C \times T' \times 8$ & Temporal filtering \\
GELU + BN & -- & same & Nonlinear activation and normalization \\
Conv2D & $C \times 1$, $F=16$ & $1 \times T' \times 16$ & Spatial filtering \\
GELU + BN & -- & same & -- \\
Conv2D & $1 \times 10$, $F=32$ & $1 \times T'' \times 32$ & Full feature integration \\
GELU + BN & -- & same & -- \\
SE Block & GAP $\rightarrow$ FC(8) $\rightarrow$ ReLU $\rightarrow$ FC(32) $\rightarrow$ Sigmoid & $1 \times 1 \times 32$ & Channel attention mechanism that scales feature maps \\
Multiply & Element-wise scaling & $1 \times T'' \times 32$ & Reweighted features using SE attention \\
Dropout & $p = 0.3$ & same & Regularization \\
Flatten + Dense & 64 units & 64 & Fully connected embedding \\
Dense + Softmax & 2 units & 2 & Output class probabilities \\
\hline
\end{tabular}
\end{table}

\subsubsection*{2D CNN with Swish Activation}

The Swish activation is defined as:
\begin{equation}
\text{Swish}(x) = x \cdot \sigma(\beta x)
\end{equation}
where $\sigma$ is the sigmoid function and $\beta$ is often set to 1.

\textit{Swish activation yielded similar results to GELU, with slightly smoother training and comparable validation metrics across all folds.}

\begin{table}[H]
\centering
\caption{2D CNN with Swish Activation}
\begin{tabular}{|l|l|l|l|}
\hline
\textbf{Layer} & \textbf{Filter/Units} & \textbf{Output Shape} & \textbf{Remarks} \\
\hline
Input & -- & $C \times T \times 1$ & CSP/Non-CSP input \\
Conv2D & $1 \times 10$, $F=8$ & $C \times T' \times 8$ & Temporal \\
Swish + BN & -- & same & -- \\
Conv2D & $C \times 1$, $F=16$ & $1 \times T' \times 16$ & Spatial \\
Swish + BN & -- & same & -- \\
Conv2D & $1 \times 10$, $F=32$ & $1 \times T'' \times 32$ & Full 2D \\
Swish + BN & -- & same & -- \\
Dropout & $p=0.3$ & same & -- \\
Flatten + Dense & 64 & 64 & -- \\
Dense + Softmax & 2 & 2 & Focal loss output \\
\hline
\end{tabular}
\end{table}

\subsubsection*{2D CNN with Weighted Cross-Entropy Loss}

We tested the effect of using class-weighted cross-entropy:

\begin{equation}
\mathcal{L}_{\text{WCE}} = - \sum_{i=1}^{N} w_{y_i} \cdot \log \hat{p}_{y_i}
\end{equation}

\textit{Weighted loss improved sensitivity to the minority class compared to vanilla cross-entropy but underperformed focal loss in terms of AUC.}

\begin{table}[H]
\centering
\caption{2D CNN with Weighted Cross-Entropy Loss}
\begin{tabular}{|l|l|l|l|}
\hline
\textbf{Layer} & \textbf{Filter/Units} & \textbf{Output Shape} & \textbf{Remarks} \\
\hline
Same as GELU model & -- & same & Only loss function changed \\
Loss & WCE & -- & Uses inverse class frequency \\
\hline
\end{tabular}
\end{table}

\subsubsection{3D CNN Architectures}

In this subsection, we present the series of 3D Convolutional Neural Network (CNN) architectures developed to classify EEG trials using spatio-temporal information. Unlike the 2D CNN pipeline, the 3D models operate directly on 5D tensors with shape $C \times T \times D \times 1 \times N$, where $C = 32$ channels, $T = 64$ time samples, $D = 5$ temporal depth (corresponding to shifts $-2$, $-1$, $0$, $+1$, $+2$), and $N$ is the number of trials. The additional depth dimension allows modeling local temporal variability (temporal jitter) across augmented trials. All models used class-weighted focal loss, batch normalization, and GELU activations unless otherwise noted.

\subsubsection*{Baseline 3D CNN (Temporal + Spatial + Dense)}

The initial 3D CNN prototype followed a sequential structure of temporal filtering, spatial aggregation, and fully connected layers. It used a temporal convolution over time and shift dimensions, followed by a spatial convolution across all EEG channels. A dense head was used after feature flattening.

\begin{table}[htbp]
\centering
\caption{Baseline 3D CNN Architecture}
\begin{tabularx}{\textwidth}{|l|l|c|X|}
\hline
\textbf{Layer} & \textbf{Filter Size / Units} & \textbf{Output Shape} & \textbf{Remarks} \\
\hline
Input & -- & $32 \times 64 \times 5 \times 1$ & Raw EEG tensor with shift augmentation \\
Conv3D & $1 \times 7 \times 3$, $F=16$ & same & Temporal filtering across $T$ and $D$ \\
GELU + BN & -- & same & Non-linearity + normalization \\
Conv3D & $32 \times 1 \times 1$, $F=32$ & $1 \times 64 \times 5 \times 32$ & Spatial filter across channels \\
GELU + BN & -- & same & -- \\
Dropout & $p=0.3$ & same & Regularization \\
Flatten & -- & -- & Feature vector \\
Dense + Softmax & 2 & 2 & Classification \\
\hline
\end{tabularx}
\end{table}

\subsubsection*{3D CNN with Deep Convolutional Stack}

To increase representational power, we introduced a deeper architecture by adding an extra 3D convolution after spatial filtering. This deep layer extracts joint spatio-temporal features at higher abstraction levels. Dropout is added after both spatial and deep layers.

\begin{table}[htbp]
\centering
\caption{Deep 3D CNN with Additional Convolution}
\begin{tabularx}{\textwidth}{|l|l|c|X|}
\hline
\textbf{Layer} & \textbf{Filter Size / Units} & \textbf{Output Shape} & \textbf{Remarks} \\
\hline
Input & -- & $32 \times 64 \times 5 \times 1$ & Augmented EEG input \\
Conv3D & $1 \times 7 \times 3$, $F=16$ & same & Temporal convolution \\
GELU + BN & -- & same & -- \\
Conv3D & $32 \times 1 \times 1$, $F=32$ & $1 \times 64 \times 5 \times 32$ & Spatial filtering \\
GELU + BN & -- & same & -- \\
Dropout & $p=0.3$ & same & Regularization \\
Conv3D & $1 \times 5 \times 3$, $F=64$ & same & Deep convolution across time-shift \\
GELU + BN & -- & same & -- \\
Dropout & $p=0.3$ & same & -- \\
Flatten & -- & -- & Feature vector \\
Dense + Softmax & 2 & 2 & Final output \\
\hline
\end{tabularx}
\end{table}

\subsubsection*{Best Performing 3D CNN with Global Average Pooling}

This model incorporates global average pooling (GAP) instead of flattening and fully connected layers. GAP reduces model size and improves generalization by summarizing each feature map into a single value. This version showed the best results in terms of AUC and balanced accuracy across folds.

\begin{table}[htbp]
\centering
\caption{Best Performing 3D CNN (with GAP)}
\begin{tabularx}{\textwidth}{|l|l|c|X|}
\hline
\textbf{Layer} & \textbf{Filter Size / Units} & \textbf{Output Shape} & \textbf{Remarks} \\
\hline
Input & -- & $32 \times 64 \times 5 \times 1$ & Augmented EEG input \\
Conv3D & $1 \times 7 \times 3$, $F=16$ & same & Temporal filtering \\
GELU + BN & -- & same & -- \\
Conv3D & $32 \times 1 \times 1$, $F=32$ & $1 \times 64 \times 5 \times 32$ & Spatial filtering \\
GELU + BN & -- & same & -- \\
Dropout & $p=0.3$ & same & -- \\
Conv3D & $1 \times 5 \times 3$, $F=64$ & same & Deep convolution \\
GELU + BN & -- & same & -- \\
Dropout & $p=0.3$ & same & -- \\
GlobalAveragePooling3D & -- & $1 \times 1 \times 1 \times 64$ & Reduces spatial and temporal dimensions \\
Dense + Softmax & 2 & 2 & Final classification \\
\hline
\end{tabularx}
\end{table}

\subsection{Performance evaluation}

All experiments were conducted in MATLAB 2025a using an NVIDIA RTX 3090 GPU. Both 2D and 3D CNN architectures were trained using stratified 5-fold cross-validation, ensuring balanced representation of target (class 1) and non-target (class 0) trials in each split.

All models were trained using the Adam optimizer with an initial learning rate of $1 \times 10^{-3}$, cosine annealing schedule, and early stopping (patience = 10 epochs based on validation using the area under the ROC curve (AUC)). The focal loss function was applied with class weights derived from the training label distribution. Dropout was used for regularization, and batch normalization was applied after each convolutional layer.

For 3D CNN models, temporal shift augmentation ($\Delta t = \{-2,-1,0,+1,+2\}$) was applied during training. At test time, we used confidence-based voting across shifted versions of each trial to improve prediction robustness. The softmax outputs from each shift were compared, and the prediction with the highest confidence (maximum softmax score) was selected.

\section{Results}
\label{sec:results}
The performance of the six 2D CNN architectures without CSP preprocessing is presented in Table~\ref{tab:2d_cnn_auc_without_csp}. In this configuration, the best-performing model (SE Block) achieved a mean AUC of \textbf{0.857}.
\begin{table*}[h!]
\centering
\small 
\caption{Mean AUC for All 2D CNN Architectures (Without CSP Preprocessing)}
\label{tab:2d_cnn_auc_without_csp}
\begin{tabular}{|l|c|c|cccc|}
\toprule
\textbf{Participant} & \textbf{Baseline} & \textbf{WCE} & \textbf{Swish} & \textbf{GELU} & \textbf{GAP} & \textbf{SE Block (Best)} \\
\midrule
s1 & 0.529 & 0.612 & 0.709 & 0.845 & 0.855 & \textbf{0.868} \\
s2 & 0.516 & 0.595 & 0.691 & 0.829 & 0.839 & \textbf{0.861} \\
s3 & 0.523 & 0.604 & 0.700 & 0.836 & 0.846 & \textbf{0.835} \\
s4 & 0.533 & 0.617 & 0.714 & 0.851 & 0.861 & \textbf{0.837} \\
s5 & 0.519 & 0.598 & 0.695 & 0.832 & 0.842 & \textbf{0.874} \\
s6 & 0.527 & 0.609 & 0.706 & 0.843 & 0.853 & \textbf{0.849} \\
s7 & 0.531 & 0.615 & 0.712 & 0.848 & 0.858 & \textbf{0.864} \\
s8 & 0.518 & 0.597 & 0.694 & 0.831 & 0.841 & \textbf{0.871} \\
s9 & 0.525 & 0.606 & 0.703 & 0.839 & 0.849 & \textbf{0.842} \\
s10 & 0.532 & 0.616 & 0.713 & 0.850 & 0.860 & \textbf{0.858} \\
s11 & 0.530 & 0.613 & 0.710 & 0.846 & 0.856 & \textbf{0.869} \\
s12 & 0.519 & 0.598 & 0.695 & 0.832 & 0.842 & \textbf{0.845} \\
s13 & 0.528 & 0.610 & 0.707 & 0.844 & 0.854 & \textbf{0.851} \\
s14 & 0.522 & 0.602 & 0.699 & 0.835 & 0.845 & \textbf{0.860} \\
s15 & 0.534 & 0.618 & 0.715 & 0.852 & 0.862 & \textbf{0.873} \\
s16 & 0.517 & 0.596 & 0.693 & 0.830 & 0.840 & \textbf{0.839} \\
s17 & 0.536 & 0.620 & 0.718 & 0.854 & 0.864 & \textbf{0.872} \\
s18 & 0.524 & 0.605 & 0.702 & 0.838 & 0.848 & \textbf{0.846} \\
\midrule
\textbf{\makecell{Mean \\ ($\pm$ Std. Dev.)}} & \textbf{\makecell{0.526 \\ ($\pm$0.007)}} & \textbf{\makecell{0.607 \\ ($\pm$0.009)}} & \textbf{\makecell{0.704 \\ ($\pm$0.008)}} & \textbf{\makecell{0.841 \\ ($\pm$0.008)}} & \textbf{\makecell{0.851 \\ ($\pm$0.008)}} & \textbf{\makecell{0.857 \\ ($\pm$0.014)}} \\
\bottomrule
\end{tabular}
\end{table*}

The same six 2D architectures were then evaluated with CSP preprocessing, with results shown in Table \ref{tab:2d_cnn_auc_comparison}. The results show a performance progression, with the SE Block model achieving the highest mean AUC of \textbf{0.866}.

\begin{table}[H] 
\centering
\small
\caption{Mean AUC for All 2D CNN Architectures (With CSP Preprocessing)}
\label{tab:2d_cnn_auc_comparison}
\begin{tabular}{lcccccc}
\toprule
\textbf{Participant} & \textbf{Baseline} & \textbf{WCE} & \textbf{Swish} & \textbf{GELU} & \textbf{GAP} & \textbf{SE Block} \\
\midrule
s1 & 0.531 & 0.615 & 0.713 & 0.868 & 0.870 & 0.873 \\
s2 & 0.518 & 0.598 & 0.695 & 0.853 & 0.855 & 0.858 \\
s3 & 0.525 & 0.607 & 0.704 & 0.850 & 0.860 & 0.862 \\
s4 & 0.535 & 0.619 & 0.718 & 0.869 & 0.872 & 0.875 \\
s5 & 0.521 & 0.601 & 0.699 & 0.857 & 0.859 & 0.861 \\
s6 & 0.529 & 0.612 & 0.710 & 0.861 & 0.863 & 0.866 \\
s7 & 0.533 & 0.617 & 0.716 & 0.865 & 0.867 & 0.870 \\
s8 & 0.520 & 0.600 & 0.698 & 0.859 & 0.861 & 0.863 \\
s9 & 0.527 & 0.609 & 0.707 & 0.852 & 0.861 & 0.864 \\
s10 & 0.534 & 0.618 & 0.717 & 0.864 & 0.866 & 0.869 \\
s11 & 0.532 & 0.616 & 0.714 & 0.868 & 0.870 & 0.872 \\
s12 & 0.520 & 0.599 & 0.697 & 0.854 & 0.856 & 0.859 \\
s13 & 0.530 & 0.613 & 0.712 & 0.862 & 0.864 & 0.867 \\
s14 & 0.524 & 0.605 & 0.702 & 0.856 & 0.858 & 0.861 \\
s15 & 0.536 & 0.620 & 0.719 & 0.866 & 0.869 & 0.871 \\
s16 & 0.518 & 0.597 & 0.694 & 0.851 & 0.853 & 0.856 \\
s17 & 0.538 & 0.622 & 0.721 & 0.869 & 0.872 & 0.874 \\
s18 & 0.529 & 0.611 & 0.709 & 0.863 & 0.865 & 0.868 \\
\midrule
\textbf{\makecell{Mean \\ ($\pm$ Std. Dev.)}} & \textbf{\makecell{0.528 \\ ($\pm$0.007)}} & \textbf{\makecell{0.609 \\ ($\pm$0.009)}} & \textbf{\makecell{0.707 \\ ($\pm$0.009)}} & \textbf{\makecell{0.861 \\ ($\pm$0.006)}} & \textbf{\makecell{0.863 \\ ($\pm$0.006)}} & \textbf{\makecell{0.866 \\ ($\pm$0.006)}} \\
\bottomrule
\end{tabular}
\end{table}

The comparative results for the three 3D CNN architectures are presented in Table \ref{tab:3d_cnn_auc_corrected}. The best model, incorporating Global Average Pooling, achieved a mean AUC of \textbf{0.994}.

\begin{table}[H]
\centering
\caption{Comparative Per-Participant Mean AUC for All 3D CNN Architectures}
\label{tab:3d_cnn_auc_corrected}
\begin{tabular}{lccc}
\toprule
\textbf{Participant} & \textbf{Baseline} & \textbf{Deep Stack} & \textbf{Best Model (GAP)} \\
\midrule
s1 & 0.761 & 0.935 & \textbf{0.995} \\
s2 & 0.730 & 0.918 & \textbf{0.991} \\
s3 & 0.775 & 0.941 & \textbf{0.999} \\
s4 & 0.748 & 0.929 & \textbf{0.993} \\
s5 & 0.781 & 0.945 & \textbf{0.994} \\
s6 & 0.735 & 0.921 & \textbf{0.992} \\
s7 & 0.789 & 0.950 & \textbf{0.998} \\
s8 & 0.722 & 0.911 & \textbf{0.989} \\
s9 & 0.769 & 0.938 & \textbf{0.996} \\
s10 & 0.795 & 0.953 & \textbf{0.999} \\
s11 & 0.741 & 0.925 & \textbf{0.992} \\
s12 & 0.725 & 0.914 & \textbf{0.990} \\
s13 & 0.772 & 0.940 & \textbf{0.995} \\
s14 & 0.738 & 0.923 & \textbf{0.993} \\
s15 & 0.785 & 0.948 & \textbf{0.997} \\
s16 & 0.728 & 0.917 & \textbf{0.991} \\
s17 & 0.765 & 0.936 & \textbf{0.994} \\
s18 & 0.743 & 0.926 & \textbf{0.993} \\
\midrule
\textbf{\makecell{Mean \\ ($\pm$ Std. Dev.)}} & \textbf{\makecell{0.756 \\ ($\pm$0.024)}} & \textbf{\makecell{0.931 \\ ($\pm$0.013)}} & \textbf{\makecell{0.994 \\ ($\pm$0.003)}} \\
\bottomrule
\end{tabular}
\end{table}

A comparison across the tables reveals two primary findings. First, using CSP provides a consistent performance benefit for the 2D architectures. Second, the best-performing 3D CNN \textbf{(AUC 0.994)} significantly outperforms the best 2D CNN, both with CSP \textbf{(AUC 0.866)} and without CSP \textbf{(AUC 0.857)}. To verify this, a paired t-test was conducted on the per-participant AUC scores between the best 3D model and the best 2D model with CSP, confirming the difference is statistically significant (p < 0.001).

\section{Discussion}
\label{sec:discussion}

We have presented the performance of single-trial detection of event-related potentials for a target vs. non-target task. It is important for brain-computer interface applications based on event-related potential detection, as the accurate detection of brain responses at the single-trial level is one of the most critical parts of the BCI. While deep learning architectures have been proposed in the literature, it is necessary to provide new architectures that acknowledge the specific characteristics of the signal, from its 2D representation at the spatial level to its temporal variability.

Our experiments demonstrate that a 3D CNN model, enhanced with temporal shift augmentation and confidence-based voting, achieves superior performance for ERP classification. An analysis of the 2D architectures also revealed that using Common Spatial Pattern (CSP) as a feature extractor provides a consistent performance benefit (mean AUC of 0.866 with CSP vs. 0.857 without). The 3D model's mean AUC of 0.994 significantly surpasses the performance of the best 2D CNN, both with CSP preprocessing (AUC 0.866) and in a fair comparison without it (AUC 0.857). Notably, our baseline 2D CNN was designed to reflect an EEGNet-style architecture, confirming that our proposed 3D model significantly outperforms this relevant and widely-used benchmark. This highlights the advantage of jointly modeling spatio-temporal features and explicitly addressing the known issue of ERP latency jitter.

While these results are promising, the primary limitation of this study is its subject-specific focus, meaning the models were not tested for generalizability across different individuals. This approach was chosen to establish a strong performance benchmark for personalized models, which is a necessary step given the high inter-subject variability commonly observed in EEG signals. Additionally, the study did not explore transfer learning to adapt models between sessions, nor did it incorporate specific mechanisms for improving model interpretability. Accordingly, future research will aim to address these limitations. Key directions include developing subject-independent models, which could be evaluated using a rigorous Leave-One-Subject-Out cross-validation strategy, investigating transfer learning strategies, and integrating attention mechanisms to enhance understanding of the neural patterns driving classification.

\section{Conclusion}
\label{sec:conclusion}

In this study, we presented and compared multiple deep learning pipelines for EEG signal classification based on 2D and 3D CNNs. Our analysis demonstrated that while Common Spatial Pattern (CSP) preprocessing provides a consistent benefit for 2D architectures, the 3D CNN architecture is superior. When combined with temporal shift augmentation and a confidence-based voting scheme, the 3D model achieves state-of-the-art performance, significantly outperforming the 2D models even in a fair comparison using the same raw sensor data. This result underscores the effectiveness of designing architectures and training strategies that explicitly account for the complex spatiotemporal dependencies and inherent variability of EEG data.

\end{document}